\documentclass{article}

\usepackage[utf8]{inputenc}
\usepackage[T1]{fontenc}
\usepackage{hyperref}
\usepackage{url}
\usepackage{booktabs}
\usepackage{amsfonts}
\usepackage{nicefrac}
\usepackage{microtype}

\usepackage{amsmath}
\usepackage{graphicx}
\usepackage{subcaption}
\usepackage{xcolor}
\usepackage{listings}
\usepackage{txfonts}
\usepackage{upgreek}

\definecolor{darkblue}{rgb}{0,0.08,0.45}
\hypersetup{
    colorlinks=true,
    linkcolor=darkblue,
    citecolor=darkblue,
    filecolor=darkblue,
    urlcolor=darkblue,
}

\definecolor{codebg}{rgb}{0.95,0.95,0.97}
\definecolor{codegreen}{rgb}{0,0.6,0}
\definecolor{codered}{rgb}{0.6,0,0}
\definecolor{codegray}{rgb}{0.5,0.5,0.5}
\definecolor{codepurple}{rgb}{0.58,0,0.82}
\definecolor{keywordcolor}{rgb}{0, 0.2, 0.7}
\definecolor{pinkcolor}{rgb}{0.75, 0.22, 0.17}

\makeatletter
\newcommand\BeraMonottfamily{
  \def\fvm@Scale{0.85}
  \fontfamily{fvm}\selectfont
}
\makeatother

\lstdefinestyle{pythonstyle}{
    language=Python,
    backgroundcolor=\color{codebg},
    commentstyle=\color{codegreen},
    keywordstyle={\color{keywordcolor}},
    stringstyle=\color{codered},
    breakatwhitespace=false,
    captionpos=b,
    keepspaces=true,
    numbers=none,
    frame=single,
    showspaces=false,
    showstringspaces=false,
    showtabs=false,
    tabsize=2,
    basicstyle=\BeraMonottfamily\small,
    breaklines=false,
    literate={-}{-}1,
}

\lstdefinestyle{bashstyle}{
    language=Bash,
    backgroundcolor=\color{codebg},
    commentstyle=\color{codegreen},
    keywordstyle={\color{keywordcolor}},
    stringstyle=\color{codered},
    breakatwhitespace=false,
    captionpos=b,
    keepspaces=true,
    numbers=none,
    frame=single,
    showspaces=false,
    showstringspaces=false,
    showtabs=false,
    tabsize=2,
    basicstyle=\BeraMonottfamily\small,
    breaklines=false,
    literate=
        {-}{-}1
        {\\}{{\color{black}{{\textbackslash}}}}1
}

\usepackage[accepted]{icml2021}

\icmltitlerunning{Tonic: A Deep Reinforcement Learning Library for Fast Prototyping and Benchmarking}

\begin{document}

\twocolumn[
\icmltitle{Tonic: A Deep Reinforcement Learning Library for \\
Fast Prototyping and Benchmarking}

\begin{icmlauthorlist}
\icmlauthor{Fabio Pardo}{imp}
\end{icmlauthorlist}

\icmlaffiliation{imp}{Robot Intelligence Lab, Imperial College London, UK}

\icmlcorrespondingauthor{Fabio Pardo}{f.pardo@imperial.ac.uk}

\icmlkeywords{Machine Learning, ICML}

\vskip 0.3in
]

\printAffiliationsAndNotice{}

\setlength{\footskip}{40pt}
\thispagestyle{plain}
\pagestyle{plain}

\begin{abstract}
Deep reinforcement learning has been one of the fastest growing fields of machine learning over the past years and numerous libraries have been open sourced to support research. However, most codebases have a steep learning curve or limited flexibility that do not satisfy a need for fast prototyping in fundamental research. This paper introduces Tonic, a Python library allowing researchers to quickly implement new ideas and measure their importance by providing: 1) general-purpose configurable modules 2) several baseline agents: A2C, TRPO, PPO, MPO, DDPG, D4PG, TD3 and SAC built with these modules 3) support for TensorFlow 2 and PyTorch 4) support for continuous-control environments from OpenAI Gym, DeepMind Control Suite and PyBullet 5) scripts to experiment in a reproducible way, plot results, and play with trained agents 6) a benchmark of the provided agents on 70 continuous-control tasks. Evaluation is performed in fair conditions with identical seeds, training and testing loops, while sharing general improvements such as non-terminal timeouts and observation normalization. Finally, to demonstrate how Tonic simplifies experimentation, a novel agent called TD4 is implemented and evaluated.

\end{abstract}

\section{Introduction}

Supported by the deep learning revolution \citep{lecun2015deep, goodfellow2016deep}, reinforcement learning (RL) \citep{sutton2018reinforcement} has grown in popularity and been at the heart of many of the recent milestones in artificial intelligence. Programs are now able to surpass the best humans at ancient board games \citep{silver2017mastering} and video games \citep{mnih2015human,vinyals2019grandmaster,berner2019dota}. While those achievements are impressive, they often rely on a number of simple fundamental research ideas that were originally developed independently such as Q-learning \citep{watkins1992q}, policy gradient \citep{sutton2000policy} or Monte Carlo tree search \citep{coulom2006efficient,kocsis2006bandit}.

\begin{figure}[t]
    \centering
    \includegraphics[width=3.5cm]{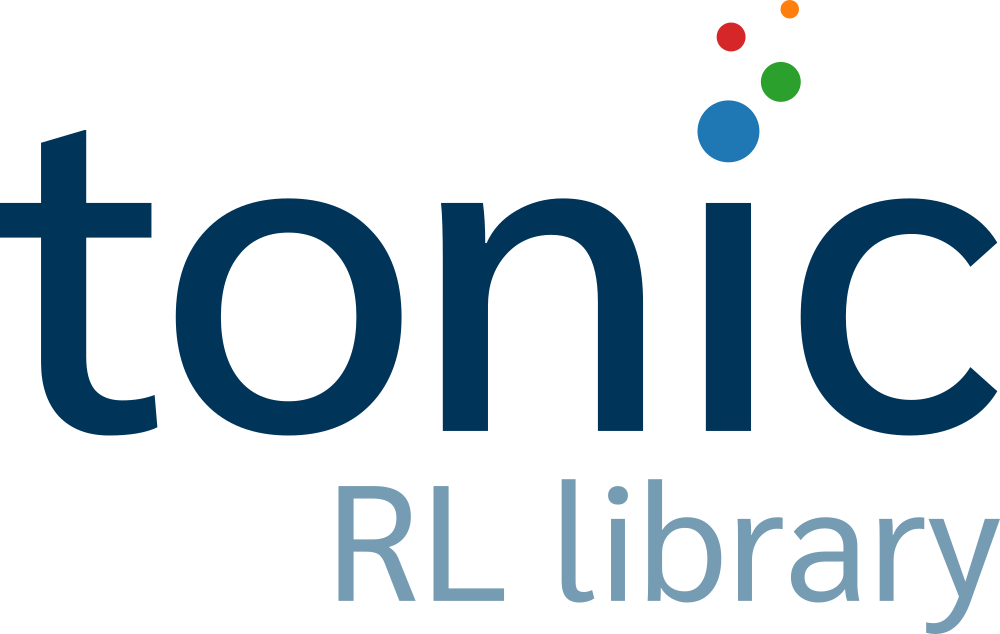}
    \caption{Tonic logo. The library and benchmark are publicly available at \href{https://github.com/fabiopardo/tonic}{github.com/fabiopardo/tonic}.}
\end{figure}

An almost systematic pattern in deep RL research is: 1) the combination of novel general purpose ideas incorporated into agents 2) performance comparison with known baseline agents on simulated environments. While writing code from scratch has formative qualities, it is usually desirable to use a simple and flexible codebase. A large number of libraries exist with diverse goals such as large scale heavily distributed training \citep[e.g.][]{liang2017rllib}, simple and pedagogical code \citep[e.g.][]{achiam2018spinning}, fundamental research in pixel-based domains \citep[e.g.][]{castro2018dopamine} or based on specific deep learning frameworks such as Keras \citep[e.g.][]{plappert2016kerasrl}, TensorFlow \citep[e.g.][]{baselines,TFAgents}, and PyTorch \mbox{\citep[e.g.][]{stooke2019rlpyt,deramo2020mushroomrl}}. While much effort has been made to build those libraries, we found that there was a need for a simple yet modular codebase designed to quickly try fundamental research ideas and evaluate them in a controlled and fair way, in particular in continuous control domains.

In this article, we introduce Tonic, a library for deep reinforcement learning research, written in Python and supporting both TensorFlow 2 \citep{abadi2016tensorflow} and PyTorch \citep{paszke2019pytorch}. Tonic includes modules such as deep learning models, replay buffers or exploration strategies. Those modules are written to be easily configured and plugged into compatible agents. Furthermore, Tonic implements a number of popular continuous control baseline agents: A2C \citep{mnih2016asynchronous}, TRPO \citep{schulman2015trust}, PPO \citep{schulman2017proximal}, MPO \citep{abdolmaleki2018maximum}, DDPG \citep{lillicrap2016continuous}, D4PG \citep{barth2018distributed}, TD3 \citep{fujimoto2018addressing} and SAC \citep{haarnoja2018soft}. Those agents are written with minimal abstractions to simplify readability and modification, emphasizing core ideas while moving other details into modules and sharing general improvements such as non-terminal timeouts \citep{pardo2018time} and observation normalization. Tonic also includes three essential scripts to 1) train and test agents in a controlled way 2) plot results against baselines and 3) play with trained policies. Finally, Tonic includes a large-scale benchmark with training logs and model weights of the baseline agents for $10$ seeds on $70$ popular environments from OpenAI Gym \citep{brockman2016openai}, DeepMind Control Suite \citep{tassa2018deepmind} and PyBullet \citep{coumans2016pybullet}, representing a large and diverse set of domains based on Box2D \citep{catto2011box2d}, MuJoCo \citep{todorov2012mujoco} and Bullet \citep{coumans2010bullet} physics engines. Table \ref{table:libraries} in the Appendix, lists a number of differences between Tonic and other popular existing RL libraries.

The paper is organized as follows: Section \ref{section:modules} presents the configuration-based philosophy underlying most of Tonic's components, Section \ref{section:trainer} presents the training pipeline, Section \ref{section:agents} describes the different agents implemented using the previously introduced modules, Section \ref{section:environments} describes the supported environments and how they are adapted to work with the Tonic agents, Section \ref{section:benchmark} presents the benchmark, some of the results and shows how a new agent called TD4 can be easily implemented and evaluated, and Section \ref{section:scripts} presents the three essential scripts provided with Tonic to simplify running experiments, interpret the results and play with the trained policies.

Before diving into the description of the library and the results, here is a minimal usage example:

\begin{lstlisting}[style=pythonstyle]
from tonic import environments, logger, Trainer
from tonic.tensorflow import agents

agent = agents.MPO()

env_fn = lambda: environments.Gym('Humanoid-v3')
env = environments.distribute(env_fn, 10, 10)
test_env = environments.distribute(env_fn)

logger.initialize('Humanoid-v3/MPO-10x10/42')

trainer = Trainer()
trainer.initialize(agent, env, test_env, 42)
trainer.run()
\end{lstlisting}

\section{Library of Modules}
\label{section:modules}

\begin{figure*}[t]
    \centering
    \includegraphics[width=\linewidth]{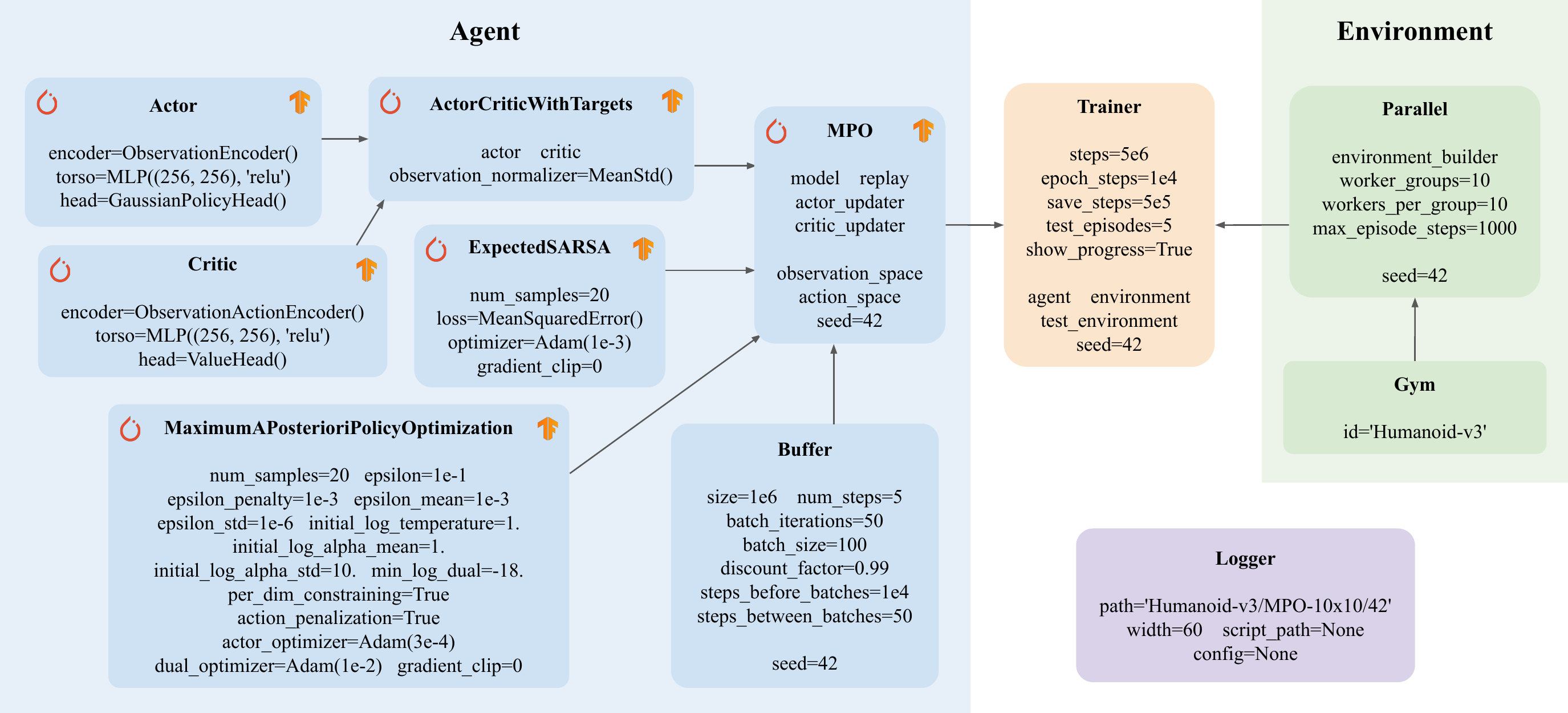}
    \caption{A hierarchy of configured modules are used to specify experiments in Tonic. Modules written for both TensorFlow 2 and PyTorch are shown with their respective logos.}
    \label{figure:modules}
\end{figure*}

Many libraries tend to put all the parameters of an experiment at the same level, calling an agent function with an environment name and all the relevant parameters. This prevents modularity and readability. Tonic tries as much as possible to move the configurable parts into modules. This has a number of advantages: 1) it clarifies the parameter targets, avoiding confusing long lists of parameters in input 2) different compatible modules with their own specific parameters can be used 3) new capabilities can be added without modifying agents.

In Tonic, the configuration of modules happens in two stages. First, when an experiment is described, agent modules are configured with general purpose parameters such as hidden layer sizes, exploration noise scale and trace decay used in $\uplambda$-returns. Then, when the environment is selected, some specific values such as the observation and action space sizes are known and the modules are finally initialized. An illustration of the hierarchy of parameterized modules corresponding to the previous code snippet is shown in \mbox{Figure \ref{figure:modules}}.

Before describing the different modules, here is a minimal usage example showing how modules can be configured and initialized:

\begin{lstlisting}[style=pythonstyle]
model = models.ActorCritic(
  actor=models.Actor(
    encoder=models.ObservationEncoder(),
    torso=models.MLP((64, 64), 'tanh'),
    head=models.DetachedScaleGaussianPolicyHead()),
  critic=models.Critic(
    encoder=models.ObservationEncoder(),
    torso=models.MLP((64, 64), 'tanh'),
    head=models.ValueHead()),
  observation_normalizer=normalizers.MeanStd())

actor_updater = updaters.ClippedRatio(
  optimizer=optimizers.Adam(3e-4, epsilon=1e-8),
  ratio_clip=0.2, kl_threshold=0.015,
  entropy_coeff=0.01, gradient_clip=40)

model.initialize(
    env.observation_space, env.action_space)
actor_updater.initialize(model)
\end{lstlisting}

\paragraph{Models} For TensorFlow 2 and PyTorch models, smaller modules are assembled. For example, an actor-critic accepts an actor and a critic network. Actors and critics are built with an encoder, a torso and a head module. An encoder processes inputs, for example concatenating observations and actions for an action-dependent critic or normalizing observations using statistics from perceived observations so far. A torso is typically a multilayer perceptron (MLP) or a recurrent network. A head produces the outputs, such as values or distributions.

\paragraph{Replays} Different kinds of replays can be used for different types of agents. For example, a traditional \lstinline[style=pythonstyle]{Buffer} can be used to randomly sample past transitions for off-policy training and a \lstinline[style=pythonstyle]{Segment} can be used to store contiguous transitions for an on-policy agent. Since those replays are configurable modules, they hold parameters like the discount factor or trace decay and are in charge of producing training batches.

\paragraph{Explorations} Different exploration strategies can be used with deterministic actors. Tonic currently includes \lstinline[style=pythonstyle]{Uniform} and \lstinline[style=pythonstyle]{Normal} for temporally-uncorrelated action noise and \lstinline[style=pythonstyle]{OrnsteinUhlenbeck} for temporally-correlated action noise.

\paragraph{Updaters} Different agents have different ways to update the parameters of their models. An updater typically takes batches of values in input, generates a loss, computes the gradient of this loss with respect to some model parameters and updates those parameters. Some updaters even create new sets of parameters used during optimization such as the dual variables in \lstinline[style=pythonstyle]{MaximumAPosterioriPolicyOptimization}.

\paragraph{Logger} The different values generated by the interaction of the agent with the environments and the values generated by the updaters are written in a \lstinline[style=bashstyle]{csv} file by the logger after each training epoch. When writting those values, outputs are also printed on the terminal in a readable table while a progress bar indicates the remaining time for the current epoch and overall training. The path to the folder containing the logs is usually of the form \lstinline[style=pythonstyle]{'environment/agent/seed/'}. When starting an experiment, the logger can also save the launch script and arguments for future reference and reloading. 

\section{Trainer}
\label{section:trainer}

\begin{figure*}
    \centering
    \includegraphics[width=0.81\linewidth]{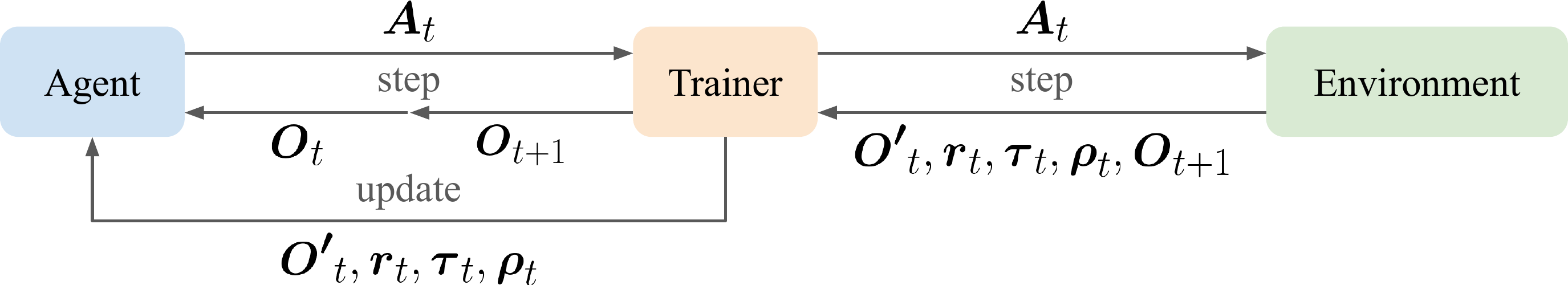}
    \caption{Synchronous training loop. At every step, a batch of observations is provided to the agent, which returns a batch of actions in output. Those actions are passed to the environment which returns a batch of transition values used to update the models: next observations, rewards, terminations and resets, plus the batch of next observations used to select the next actions.}
    \label{figure:loop}
\end{figure*}

The \lstinline[style=pythonstyle]{tonic.Trainer} module is in charge of the training loop in Tonic. It takes care of the communication between the agent and the environment, testing the agent on the test environment, logging data via the logger and periodically saving the model parameters in checkpoints for future reload.

Distributed training has been shown to greatly accelerate the training of RL agents with respect to wall clock time \citep{mnih2016asynchronous,espeholt2018impala}. Instead of interacting with a single environment at a time, the agent interacts with a set of differently seeded copies of the environment to diversify experience and increase throughput. For simplicity and to ensure reproducibility, Tonic uses a synchronous training loop illustrated in Figure \ref{figure:loop}. At training step $t$, the trainer first sends a tensor $\boldsymbol{O}_{t}$ of observations to the agent via the \lstinline[style=pythonstyle]{agent.step} function which returns a tensor $\boldsymbol{A}_{t}$ of actions and keeps track of some values such as $\boldsymbol{O}_{t}$ or the log probabilities of the actions. The actions are transmitted to the environment module via the \lstinline[style=pythonstyle]{environment.step} function which returns multiple values. First, the ones describing the current transitions caused by the actions $\boldsymbol{A}_{t}$: the tensor $\boldsymbol{O'}_{t}$ of next observations, the vectors $\boldsymbol{r}_{t}$ of rewards, $\boldsymbol{\tau}_{t}$ of terminations and $\boldsymbol{\rho}_{t}$ of resets. The terminations indicate true environmental terminations, the ones caused for example by falling on the floor in a locomotion task or reaching a target state. Agents can use those to know when bootstrapping is possible. The resets vector signals the end of episodes, from terminations and timeouts and can be used by agents to know the boundaries of episodes, for example for $\uplambda$-return calculations. When using non-terminal timeouts, partial-episode bootstrapping \citep{pardo2018time} is used to bootstrap from the values in $\boldsymbol{O'}_{t}$ and it is therefore important to know that a reset happened without an environmental termination. When an environment resets, a new observation is generated and has to be used to select the next action, therefore, the environment also returns a tensor $\boldsymbol{O}_{t+1}$ of observations to use next. For a sub-environment $i$, $\boldsymbol{o'}^i_{t} = \boldsymbol{o}^i_{t+1}$ if $\rho^i_t =$ \lstinline[style=pythonstyle]{False}. Finally, the transition values are given to the actor via the \lstinline[style=pythonstyle]{actor.update} function which takes care of registering the transitions in a replay and performing updates, while the new observations $\boldsymbol{O}_{t+1}$ are used to generate the new actions $\boldsymbol{A}_{t+1}$ at the next step.

\section{Agents}
\label{section:agents}

A number of reinforcement learning agents have been proposed over the years. Tonic contains 8 popular baseline agents, some are simple and foundational while others are more complicated state of the art algorithms.

\paragraph{Basic agents} Especially useful for debugging, the simple non-parametric agents are \lstinline[style=pythonstyle]{NormalRandom}, \lstinline[style=pythonstyle]{UniformRandom}, \lstinline[style=pythonstyle]{OrnsteinUhlenbeck}, and \lstinline[style=pythonstyle]{Constant}.

\paragraph{Advantage Actor-Critic (A2C)} This agent, also called Vanilla Policy Gradient (VPG) in some libraries, uses advantages from $\uplambda$-returns and a learned value function to update a stochastic policy via policy gradient \citep{sutton2000policy,schulman2016high,mnih2016asynchronous}. It is stable but learns slowly because it can use the latest collected transitions only once to update its actor.

\paragraph{Trust Region Policy Optimization (TRPO)} This agent uses a conjugate gradient optimizer to take a large update step of policy gradient while satisfying a KL constraint between the new and previous policies \citep{schulman2015trust}.

\paragraph{Proximal Policy Optimization (PPO)} This agent approximates TRPO by using clipped ratios between the old policy which generated the latest transitions and the currently optimized policy \citep{schulman2017proximal}.

\paragraph{Maximum a Posteriori Policy Optimisation (MPO)} This agent uses a complex relative-entropy objective taking advantage of the duality between control and estimation \citep{abdolmaleki2018maximum}. It can be very powerful if carefully tuned but its complexity made it very challenging to implement and Acme's code \citep{hoffman2020acme} was the only reliable source when Tonic was created.

\paragraph{Deep Deterministic Policy Gradient (DDPG)} This agent uses a deterministic actor trained via deterministic policy gradient \citep{silver2014deterministic,lillicrap2016continuous}. It is data-efficient because it learns off-policy an approximation to the optimal value function used to locally optimize the actor.

\paragraph{Distributed Distributional Deep Deterministic Policy Gradient (D4PG)} This agent uses a distributional critic head, n-step returns and prioritized experience replay \citep{barth2018distributed}. Tonic does not currently include a prioritized replay buffer but as pointed in the original paper, this is a less critical component and can lead to unstable updates.

\paragraph{Twin Delayed Deep Deterministic Policy Gradient (TD3)} This agent stabilizes DDPG using a pair of critics, action noise in the target actor and a delay to update the actor network less often \citep{fujimoto2018addressing}.

\paragraph{Soft Actor-Critic (SAC)} This agent uses an entropy based reward augmentation, a squashed Gaussian policy and a pair of critics \citep{haarnoja2018soft}.

\section{Environments}
\label{section:environments}

\begin{figure*}[t]
    \begin{flushright}
    \includegraphics[width=0.19\linewidth]{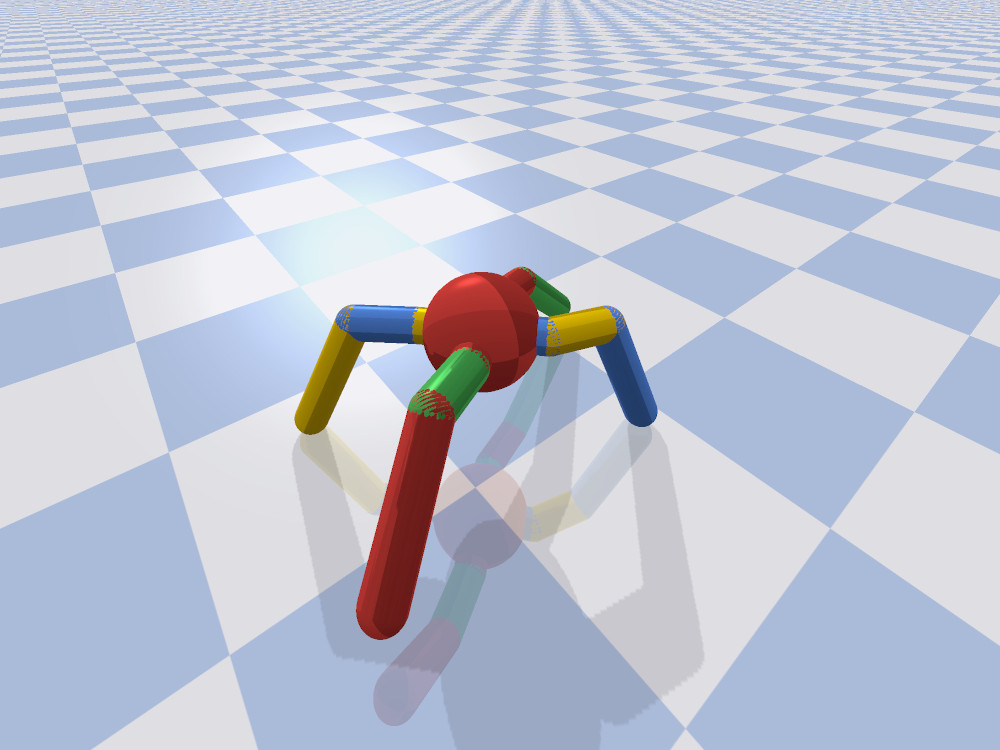}
    \includegraphics[width=0.19\linewidth]{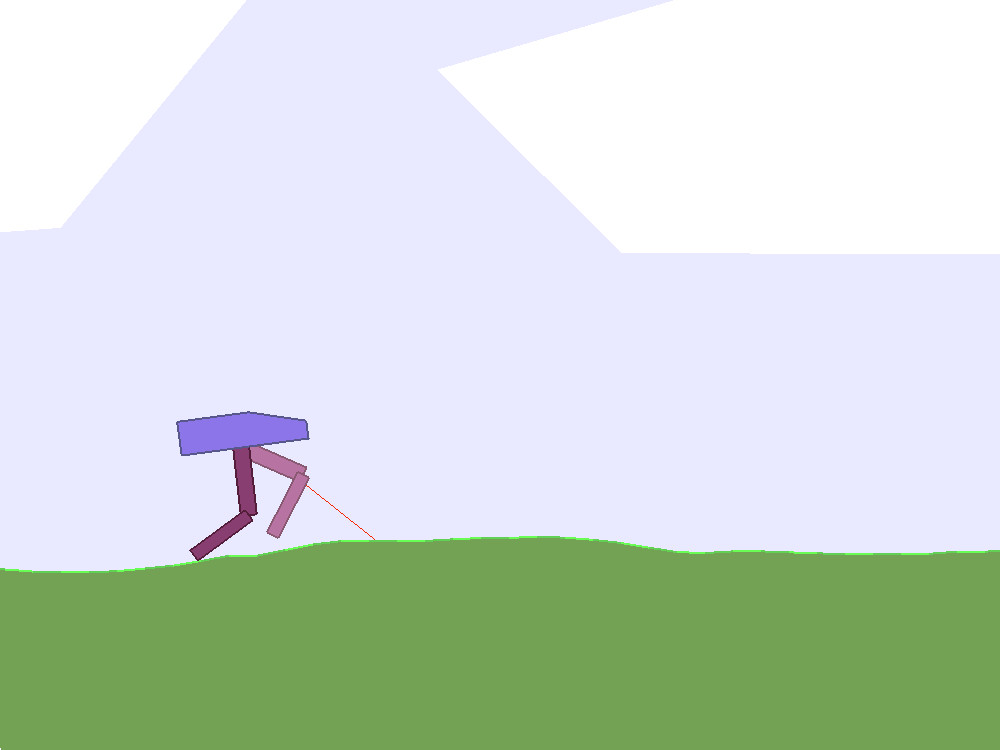}
    \includegraphics[width=0.19\linewidth]{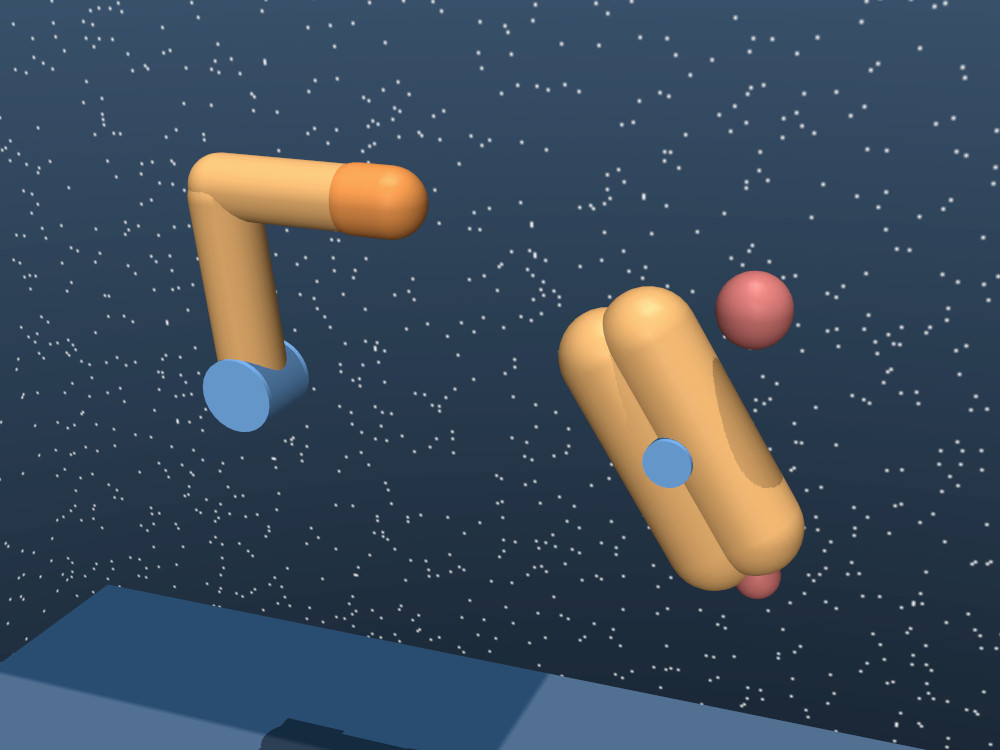}
    \includegraphics[width=0.19\linewidth]{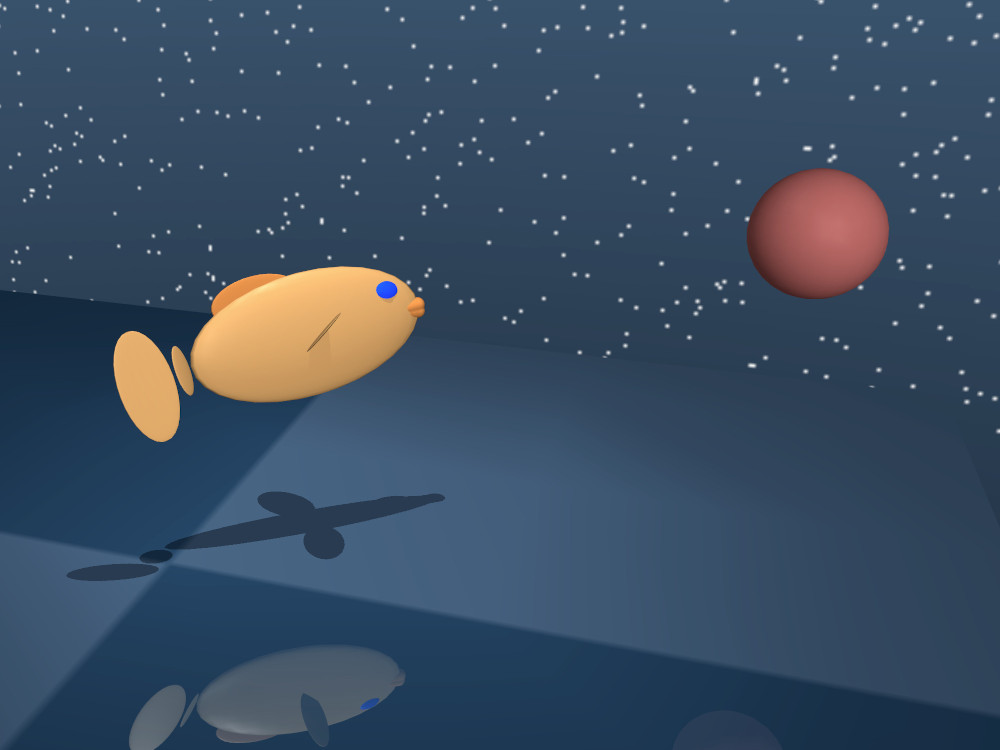}
    \includegraphics[width=0.19\linewidth]{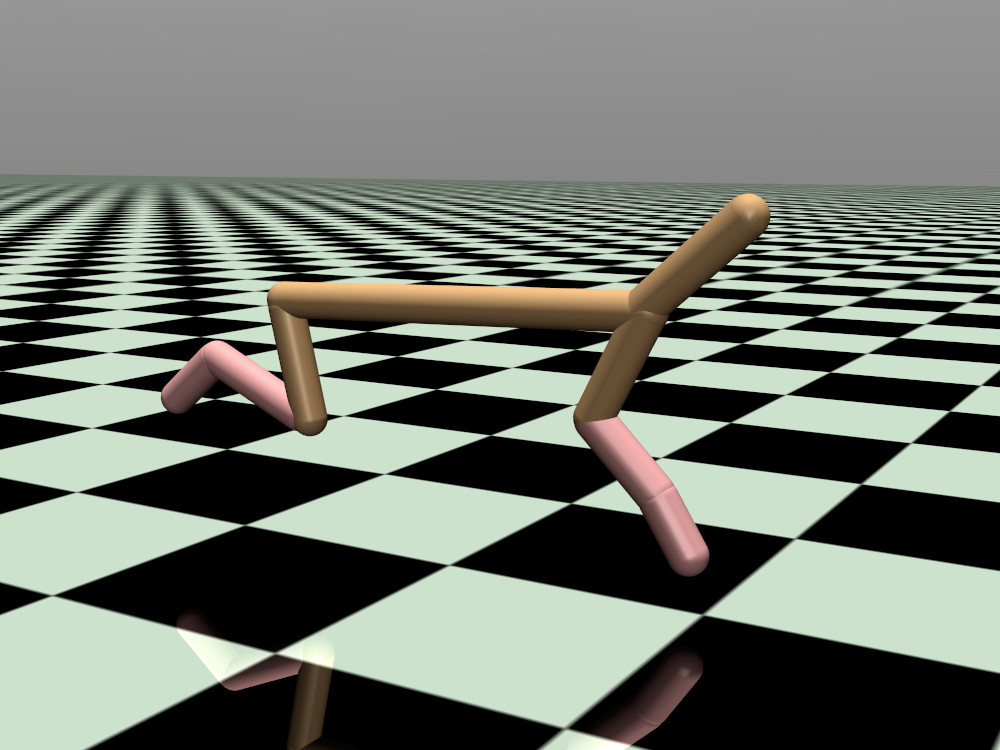}
    \includegraphics[width=0.19\linewidth]{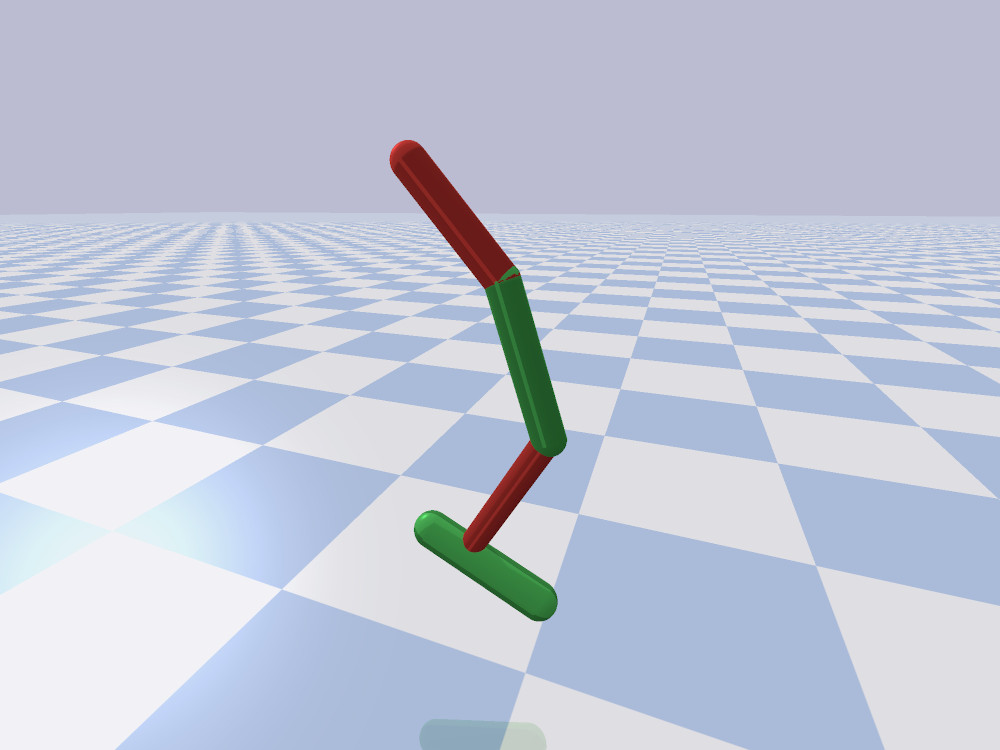}
    \includegraphics[width=0.19\linewidth]{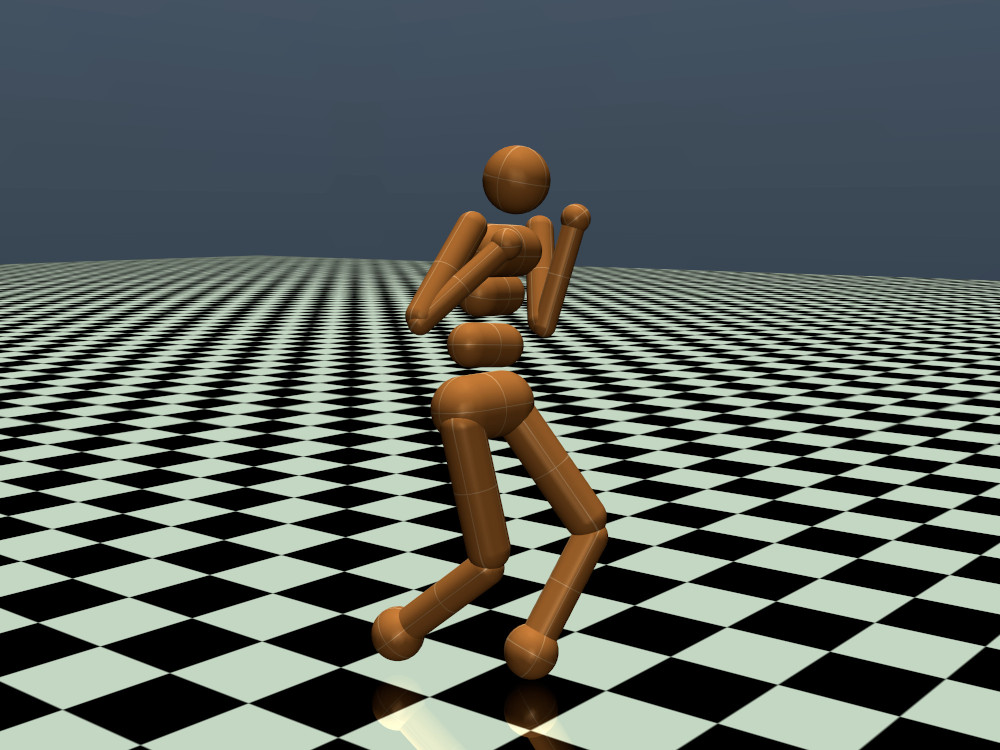}
    \includegraphics[width=0.19\linewidth]{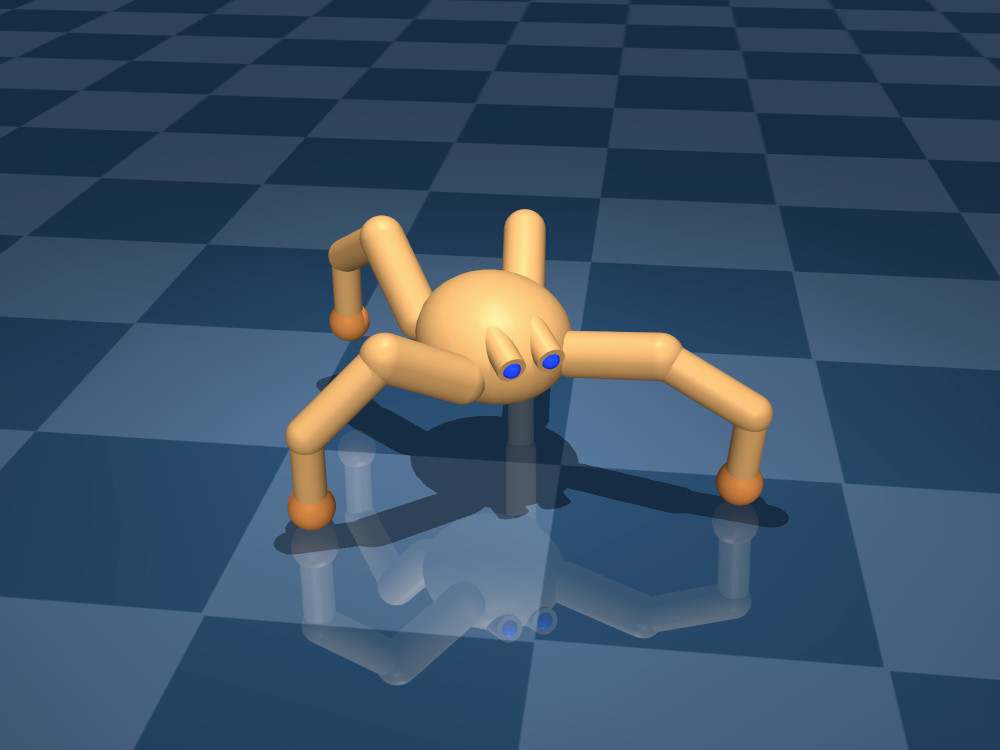}
    \includegraphics[width=0.19\linewidth]{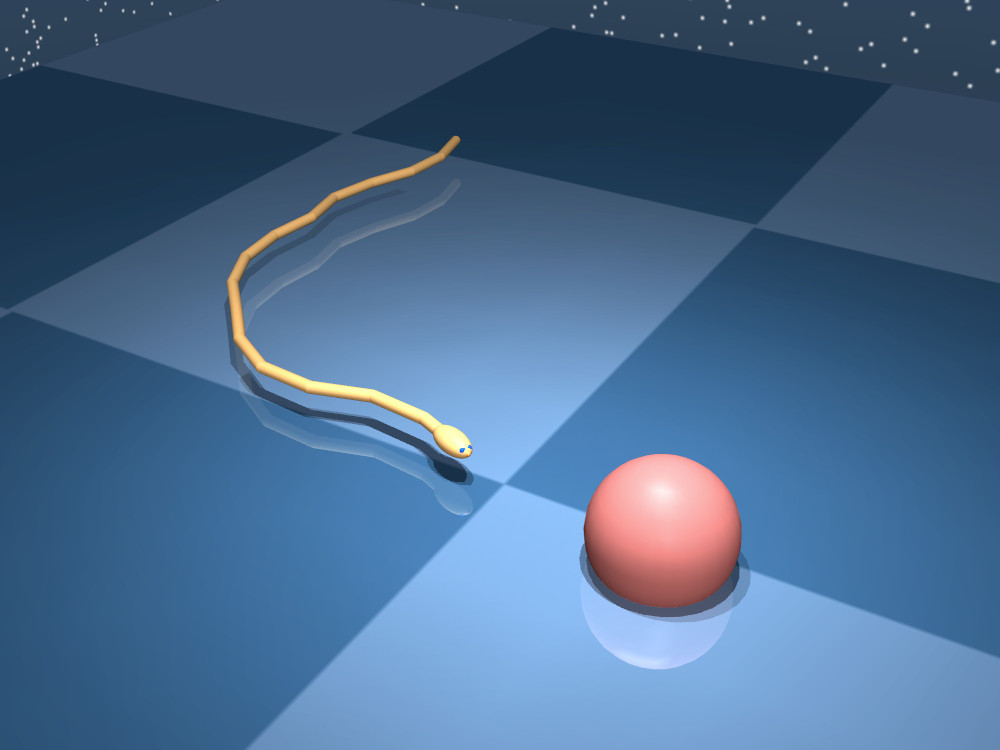}
    \includegraphics[width=0.19\linewidth]{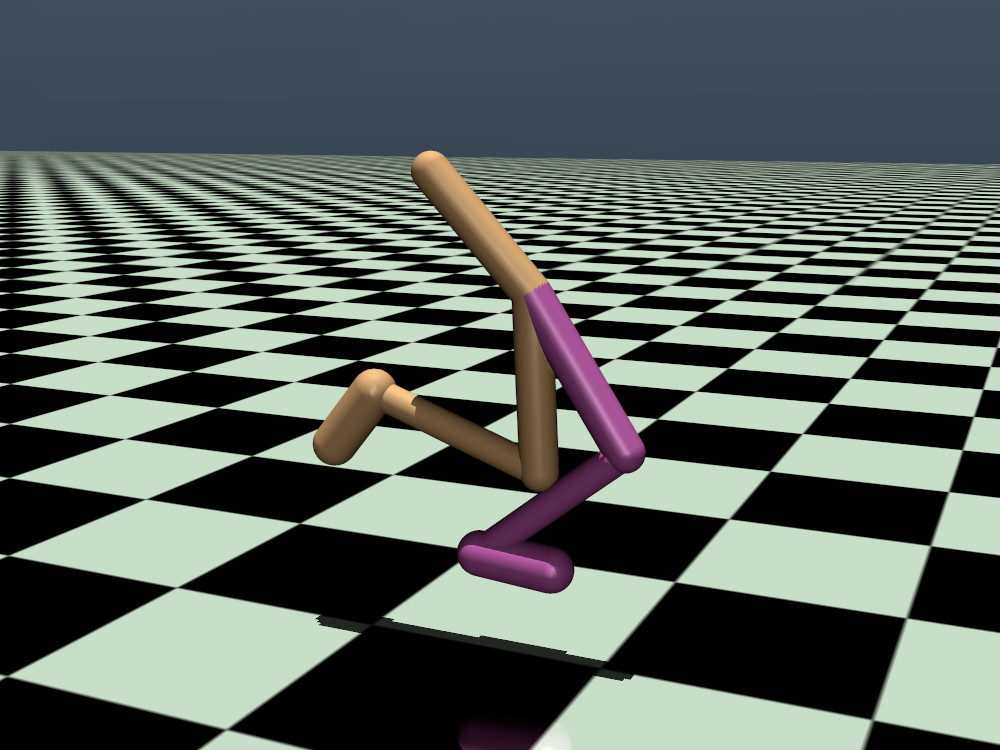}
    \end{flushright}
    \centering
    \includegraphics[width=\linewidth]{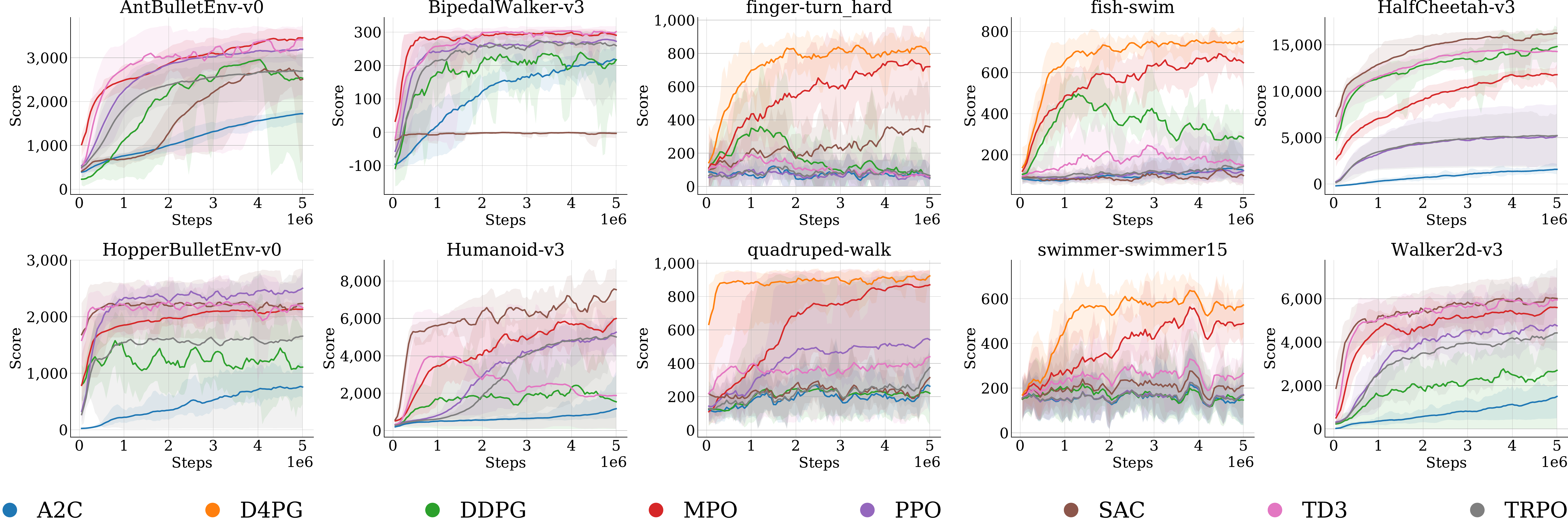}
    \caption{Subset of the benchmark results. For each agent, 5 test episodes are collected after each training epoch and averaged. The solid lines represent the average over 10 runs for each agent. The [minimum, maximum] range is shown with transparent areas and a sliding window of size 5 is used for smoothing. A large palette of environments are represented across the supported domains. The best performing agents are mostly TD3, SAC, MPO and D4PG (Control Suite tasks) but significant variations exist for each agent.}
    \label{figure:results}
\end{figure*}

\paragraph{OpenAI Gym, PyBullet and DeepMind Control Suite}
Tonic includes builders for continuous-control environments from OpenAI Gym \citep{brockman2016openai}, DeepMind Control Suite \citep{tassa2018deepmind} and PyBullet \citep{coumans2016pybullet}, representing a large and diverse set of domains based on Box2D \citep{catto2011box2d}, MuJoCo \citep{todorov2012mujoco} and Bullet \citep{coumans2010bullet} physics engines. For simplicity and to match Gym and PyBullet environments, dictionary observations are flattened and concatenated in vectors.

\paragraph{Non-terminal timeouts}
All of these environments are wrapped to enable the synchronous interaction described in Section \ref{section:trainer}. The \lstinline[style=pythonstyle]{TimeLimit} wrapper is removed from the Gym and PyBullet environments while in the case of Control Suite environments, task terminations are detected from \lstinline[style=pythonstyle]{task.get_termination(physics)}. Moreover, when \lstinline[style=pythonstyle]{terminal_timeouts} is set to \lstinline[style=pythonstyle]{True}, it is recommended to also set \lstinline[style=pythonstyle]{time_feature} to \lstinline[style=pythonstyle]{True} to use a \lstinline[style=pythonstyle]{tonic.environments.TimeFeature} wrapper, adding a representation of the remaining time in observation. This is known as time-awareness \citep{pardo2018time} and allows environments to stay Markovian.

\paragraph{Action scaling}
Agents are all expected to act in a $[-1, 1]^d$ action space where $d$ is the number of dimensions. This facilitates action noise scaling and learning for agents relying on deterministic policies. Environments use a \lstinline[style=pythonstyle]{tonic.environments.ActionRescaler} wrapper by default.

\paragraph{Distributed training}
Finally, for distributed training, the set of environment copies is maintained in parallel groups of sequential workers. Each parallel group is allocated to a different process and communication is done via pipes. Since this communication method adds some time overhead, using multiple sequential environments in each group can increase throughput.

\section{Benchmark}
\label{section:benchmark}

\begin{figure*}[t]
    \includegraphics[width=\linewidth]{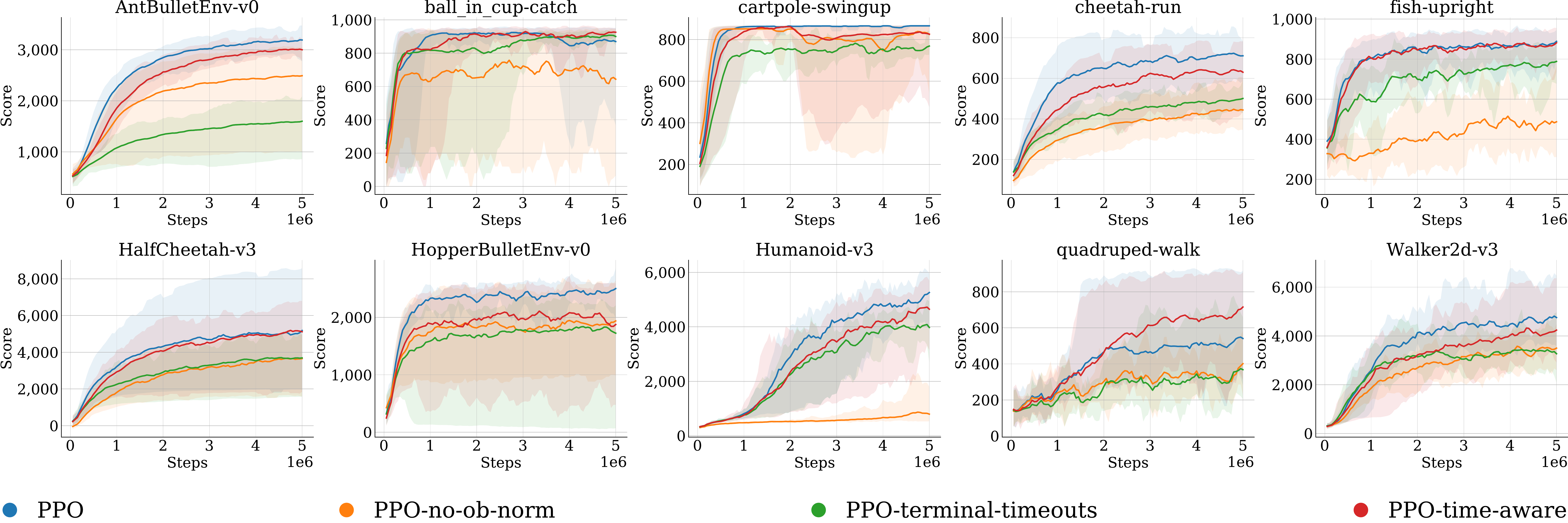}
    \caption{Efficiency of observation normalization and non-terminal timeouts. PPO uses the default configuration with observation normalization and non-terminal timeouts. PPO-ob-norm is identical but without observation-normalization. PPO-terminal-timeouts is identical to PPO but with environmental termination at timeouts as is the case originally for the supported environments. PPO-time-aware is identical to PPO-terminal-timeouts but adds the remaining time as an observed feature. Observation normalization seems to accelerate learning. Non-terminal timeouts are best while time features are needed to account for terminal timeouts.}
    \label{figure:ablations}
\end{figure*}

When evaluating novel ideas in the literature, it is sometimes difficult to measure the significance of results as baselines can be poorly tuned or evaluated in unfair conditions. Benchmarks of popular RL agents on popular RL environments \citep[e.g.][]{duan2016benchmarking,cleanrl} evaluated in identical conditions are essential to provide reliable lower bounds in fundamental research.

\paragraph{Methods}
Tonic contains a large-scale benchmark of the 8 provided deep RL agents on  70 popular continuous-control tasks: 17 from OpenAI Gym (2 classic control, 3 Box2D, 12 MuJoCo), 10 from PyBullet and 43 from the benchmark subset in DeepMind Control Suite. The exact same 10 seeds (0, 1, 2, ..., 9) are used for all agents with default parameters on all environments with single-worker training (not distributed). D4PG was only run on DeepMind Control Suite environments because known reward boundaries are required for distributional value functions. Therefore, the total number of runs contained in the benchmark is $70 \times 10 \times 7 + 43 \times 10 = 5330$. These runs were all generated with \lstinline[style=pythonstyle]{tonic.tensorflow} which was significantly faster with off-policy agents than \lstinline[style=pythonstyle]{tonic.torch}. This difference could be due to a more efficient graph tracing mechanism provided by TensorFlow's \lstinline[style=pythonstyle]{tf.function} decorator. A speed comparison is provided in Appendix Figure \ref{figure:tf_vs_pt}. The runs were started by only specifying the environment, the agent and the seed, without any other argument. This means that all the hyperparameters used are the default ones.

\paragraph{Results}
Some of the results can be seen in Figure \ref{figure:results} while the full benchmark plots can be found in Appendix Figure \ref{figure:benchmark}. Environments from OpenAI Gym (names starting with an uppercase) and PyBullet (names with ``PybulletEnv'') are mostly ``solved'' with best agents getting scores similar to the best performances reported in the literature. However, many environments from DeepMind Control Suite seem much harder to learn and most results reported for those environments can be found in the literature with distributed training and many more training steps. Nevertheless, it is important to note that better hyper parameters could certainly be found for those agents, and especially better ones for each environment specifically.

\paragraph{Comparison to Spinning Up in Deep RL}
To prove that the results for A2C, TRPO, PPO, DDPG, TD3 and SAC can be used as valid baselines, another benchmark was generated with TensorFlow 1 implementations of those agents from Spinning Up in Deep RL \citep{achiam2018spinning}. The library was slightly modified to use a test environment for each agent, a frequency and number of test episodes and seeds identical to the ones used in the Tonic benchmark and VPG was renamed A2C. Results on the original 5 environments used in the benchmark of this library can be found in Appendix Figure \ref{figure:spinup}. The results from Spinning Up in Deep RL are compatible with the ones found on the website. The agents from Tonic perform significantly better on four of the five environments. This difference can be explained by some of the improvements in Tonic, such as observation normalization, non-terminal timeouts and action-scaling, even though Spinning Up in Deep RL partially implements non-terminal timeouts for the off-policy methods by ignoring environmental terminations at timeout.

\paragraph{Ablations and variants}
To measure the effectiveness of non-terminal timeouts and observation normalization in particular, PPO was trained with different variants. Results shown in Figure \ref{figure:ablations} demonstrate that those improvements indeed improve performance of PPO. Finally, supplementary experiments validated the effectiveness of these improvements on the other agents and environments.

\begin{figure*}[t]
    \includegraphics[width=\linewidth]{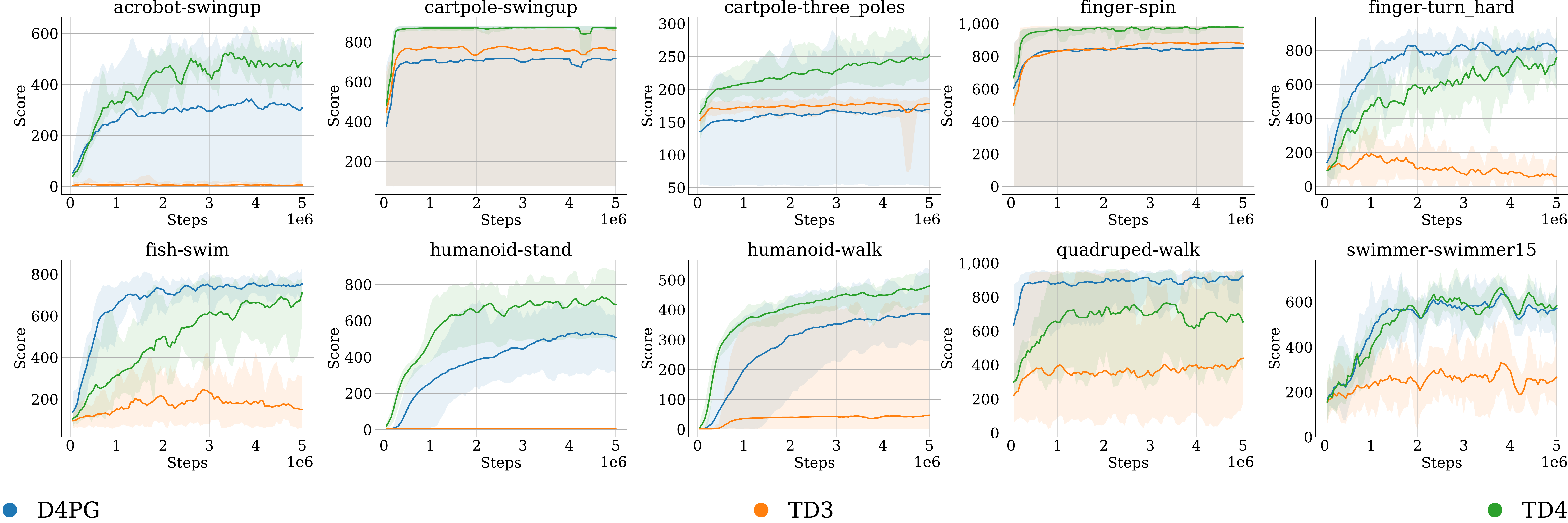}
    \caption{Evaluation of the proposed TD4 agent. The performance is significantly better on 6 of the 10 tasks, demonstrating that the agent successfully combines features from D4PG and TD3.}
    \label{figure:td4}
\end{figure*}

\paragraph{Prototyping and benchmarking a novel agent}
To demonstrate how Tonic can accelerate the development and the evaluation of ideas, a new agent called TD4, combining features of TD3 and D4PG is proposed. The source code consisting of 93 clean lines of code can be found in Appendix Section \ref{section:td4_code}. It contains three elements. The first one is a new model based on \lstinline[style=pythonstyle]{ActorTwinCriticWithTargets} with a \lstinline[style=pythonstyle]{DistributionalValueHead} for the twin critic. The second element is a new updater similar to \lstinline[style=pythonstyle]{TwinCriticDeterministicQLearning} but adapted to use a pair of critics. The third element is the agent itself, based on \lstinline[style=pythonstyle]{TD3} and \lstinline[style=pythonstyle]{D4PG} and using the two previous elements. The command lines needed to train the proposed agent and directly compare its performance to D4PG and TD3 can be found in \ref{section:td4_commands}. The results on 10 tasks are shown in Figure \ref{figure:td4} and demonstrate that TD4 is an excellent agent combining advantages from TD3 and D4PG and was particularly simple to implement and evaluate.

\section{Scripts}
\label{section:scripts}

The modules and agents described above can easily be used in a standalone experiment Python script or integrated in another codebase. However, for convenience, Tonic includes three essential scripts to take care of the most important things: training, plotting and playing.

\paragraph{tonic.train} A script used to launch training experiments. Since any agent could be configured with any compatible modules and launched on any configured environment, a simple list of parsed parameters would not give enough flexibility. Therefore, Tonic uses the interpreted nature of the Python language to directly evaluate Python snippets describing the agent, the environment and the trainer configurations. The script saves the experiment script and arguments and automatically configures the logger to use a path of the form \lstinline[style=pythonstyle]{'environment/agent/seed/'} which will be recognized by the two other scripts. Usage example:

\begin{lstlisting}[style=bashstyle]
python3 -m tonic.train \
--header "import tonic.torch" \
--agent "tonic.torch.agents.PPO()" \
--environment "tonic.environments.Gym('Ant-v3')" \
--seed 0
\end{lstlisting}

\paragraph{tonic.plot} A script to load and display results from multiple experiments together. The script expects a list of \lstinline[style=bashstyle]{csv} or \lstinline[style=bashstyle]{pkl} files to load data from. Regular expressions like \lstinline[style=bashstyle]{BipedalWalker-v3/PPO-X/0}, \lstinline[style=bashstyle]|BipedalWalker-v3/{PPO*,DDPG*}| or \lstinline[style=bashstyle]{*Bullet*} can be used to point to different sets of logs. Multiple sub-figures are generated, one per environment, aggregating results of agents across runs. The script can be configured in many ways. For example, the figure can be saved in different file formats such as \lstinline[style=bashstyle]{PDF} and \lstinline[style=bashstyle]{PNG}. A non-GUI \lstinline[style=bashstyle]{backend} such as \lstinline[style=bashstyle]{agg} can be used. If the \lstinline[style=bashstyle]{seconds} argument is used, plotting is performed regularly in real time. The \lstinline[style=bashstyle]{baselines} argument can be used to load logs from the benchmark saved in the \lstinline[style=bashstyle]{/data/logs} folder at the root of Tonic. For example, \lstinline[style=bashstyle]{--baselines all} uses all agents while \lstinline[style=bashstyle]{--baselines A2C PPO TRPO} will use logs from A2C, PPO and TRPO. Different parameters allow the user to customize the x and y axes, change the smoothing window size, specify the type of interval shown, display individual runs, select the minimum and maximum values of the x axis, and do many other things. Finally, the legend is shown at the bottom of the figure, regrouping all agents across environments with a mechanism to automatically detect the ideal number of legend columns to use. Usage example:

\begin{lstlisting}[style=bashstyle]
python3 -m tonic.plot --path Ant-v3 --baselines all
\end{lstlisting}

\paragraph{tonic.play} A script to reinstantiate the environment and agent from an experiment folder, reloading weights from a checkpoint and rendering the policy acting in the environment. The path to the experiment must be specified and a particular checkpoint can be chosen. While rendering the policy interacting with the environment, the episode lengths, scores, min and max rewards are printed on the terminal. Gym environments are simply rendered while PyBullet and DeepMind Control Suite viewers allow users to add perturbations to the bodies in the simulation. Usage example:

\begin{lstlisting}[style=bashstyle]
python3 -m tonic.play --path BipedalWalker-v3/PPO/0
\end{lstlisting}

\paragraph{Adding new modules, agents and environments} When using \lstinline[style=bashstyle]{tonic.train}, new components can be added using the \lstinline[style=bashstyle]{header} field which is evaluated first. For example, if the components are installed or accessible from the current directory, they can directly be imported. If necessary, the path to the required files can also be added to the \lstinline[style=bashstyle]{sys.path} before importing them. Finally, for OpenAI Gym, PyBullet and DeepMind Control Suite environments, it is recommended to register new tasks at import time in the packages themselves. For example the \lstinline[style=bashstyle]{__init__.py} at the root of a package containing new tasks could use \lstinline[style=pythonstyle]{register(id='Cat-v0', entry_point=CatEnv)} for Gym and PyBullet, and \lstinline[style=pythonstyle]{suite._DOMAINS['cat'] = cat_tasks} for Control Suite. Usage example:

\begin{lstlisting}[style=bashstyle]
python3 -m tonic.train \
--header "import animal_envs, tonic.tensorflow" \
--environment "tonic.environments.Gym('Cat-v0')" \
--agent "tonic.tensorflow.agents.TD3()" \
--seed 0
\end{lstlisting}

\section{Conclusion and Future Work}

This paper introduced Tonic, a library designed for fast prototyping and benchmarking of deep RL algorithms. It contains a number of configurable modules, agents, supported environments, three essential scripts and a large-scale continuous-control benchmark. Future work will include support for discrete action spaces and pixel-based observations, better handling of dictionary-based observations, benchmark results with improved hyper-parameters, new modules and agents. In particular, some of the new agents will rely on discretization of continuous-action spaces as this alternative has proven to be competitive with continuous-control methods \citep{metz2017discrete,tavakoli2018action,van2020q}. Hopefully researchers will use Tonic, contribute to it and will find easier to release the source code of their papers.

\section*{Acknowledgements}

We thank Arash Tavakoli, Nemanja Rakicevic and Digby Chappell for comments on the manuscript. Tonic was inspired by a number of other deep RL codebases. In particular, we acknowledge OpenAI Baselines, Spinning Up in Deep RL and Acme. The research presented in this paper has been supported by Dyson Technology Ltd. 

\bibliography{main}
\bibliographystyle{icml2021}

\clearpage
\onecolumn

\appendix

\section{Benchmark Results}

\begin{figure}[h!]
    \centering
    \includegraphics[width=\linewidth]{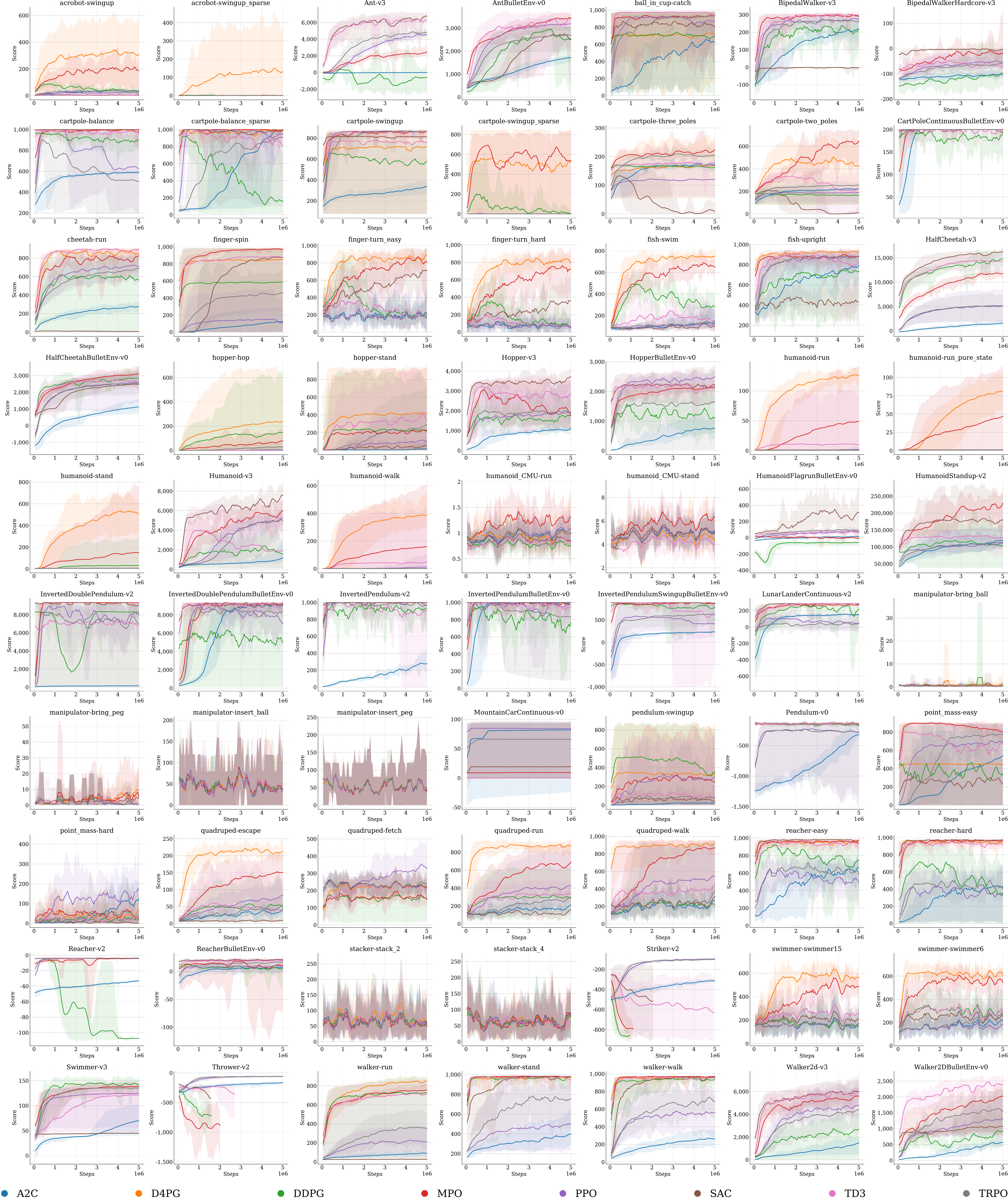}
    \caption{Full benchmark results on 70 tasks, for 5 million time steps over 10 seeds. Intervals indicate minimum and maximum values. Curves are smoothed with a sliding window of size 5. A few runs diverged catastrophically, especially on Striker-v2 and Thrower-v2.}
    \label{figure:benchmark}
\end{figure}

\clearpage
\section{Comparison with Spinning Up in Deep RL}

\begin{figure}[h!]
    \centering
    \includegraphics[width=\linewidth]{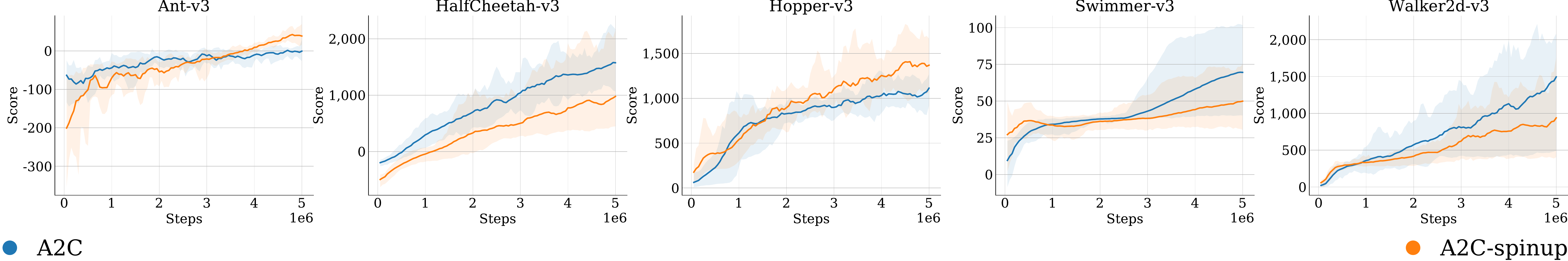}\\
    \vspace{0.2cm}
    \includegraphics[width=\linewidth]{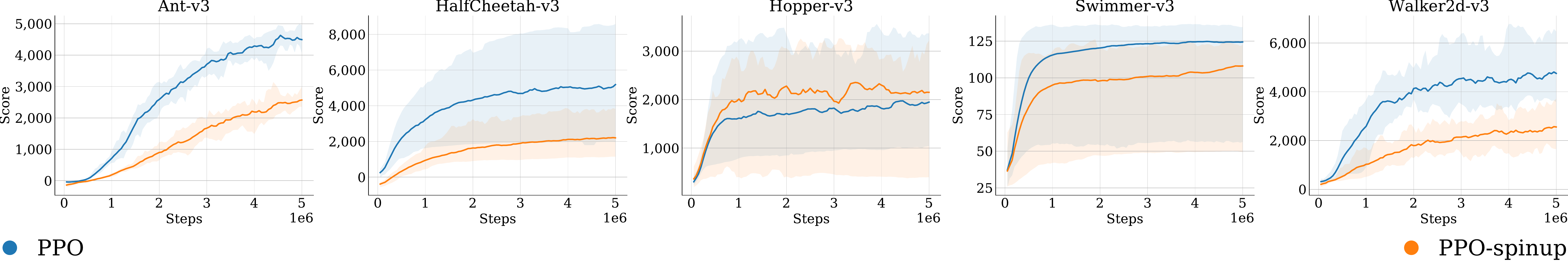}\\
    \vspace{0.2cm}
    \includegraphics[width=\linewidth]{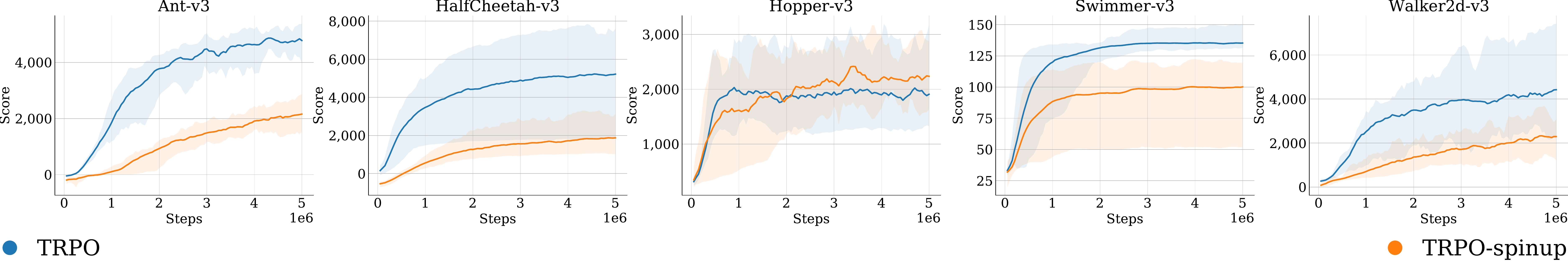}\\
    \vspace{0.2cm}
    \includegraphics[width=\linewidth]{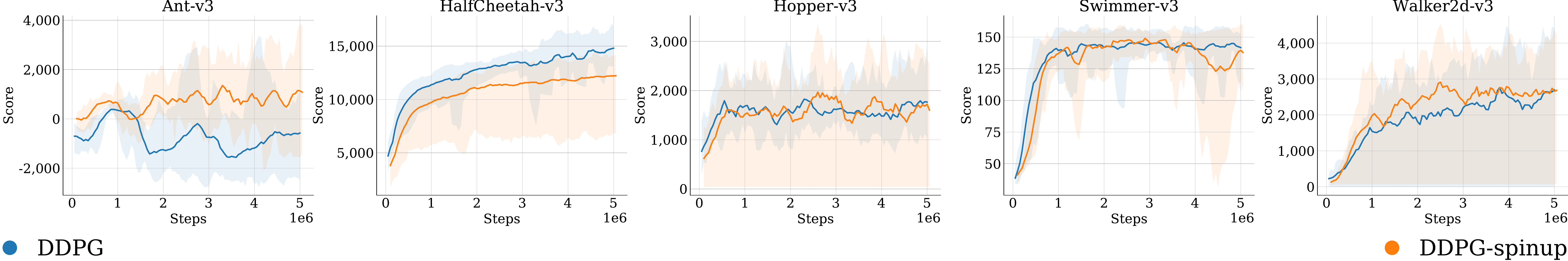}\\
    \vspace{0.2cm}
    \includegraphics[width=\linewidth]{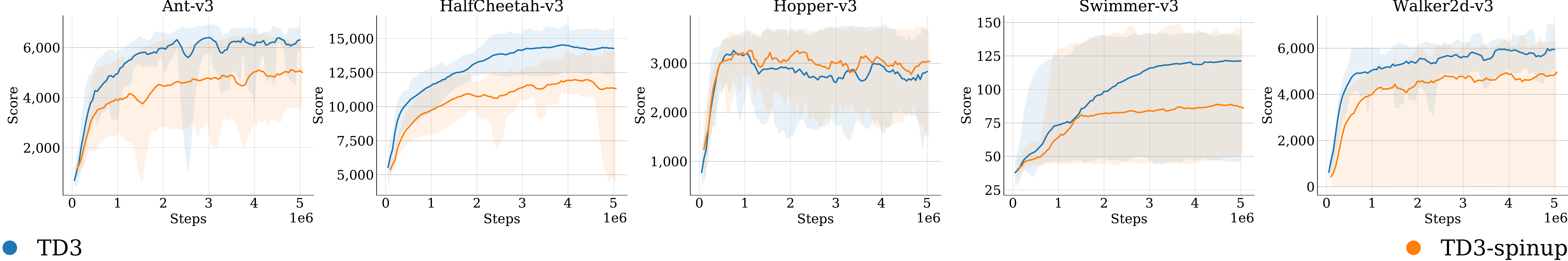}\\
    \vspace{0.2cm}
    \includegraphics[width=\linewidth]{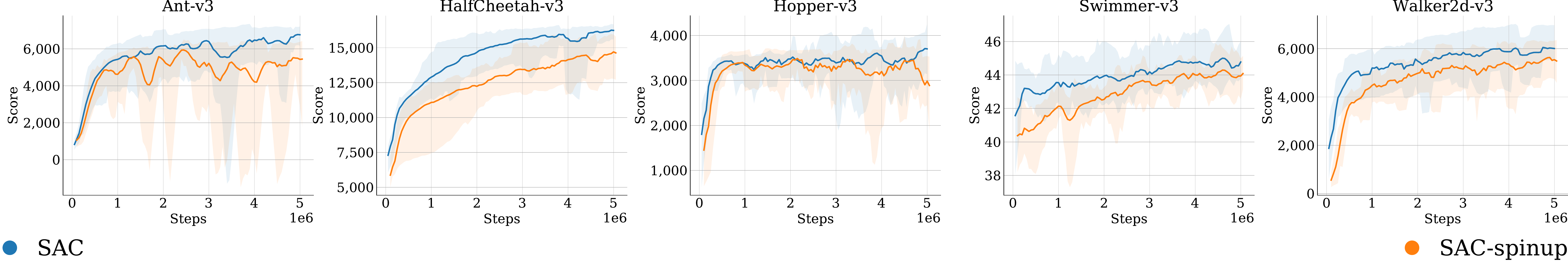}
    \caption{Comparison with Spinning Up in Deep RL using the same training, evaluation and agent parameters.}
    \label{figure:spinup}
\end{figure}

\clearpage
\section{TensorFlow 2 vs PyTorch}

\begin{figure}[h!]
    \centering
    \includegraphics[height=6cm]{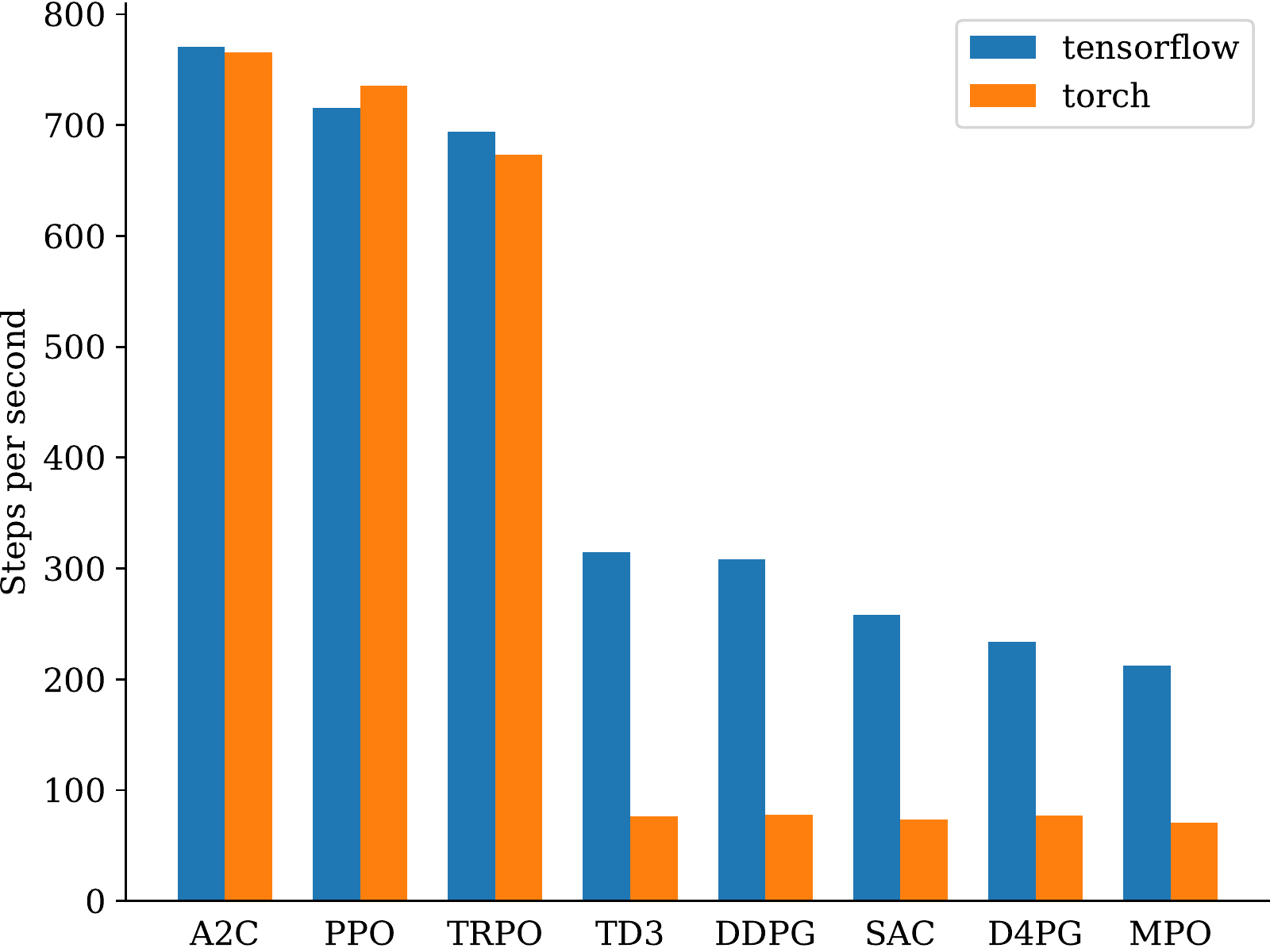}
    \includegraphics[height=6cm]{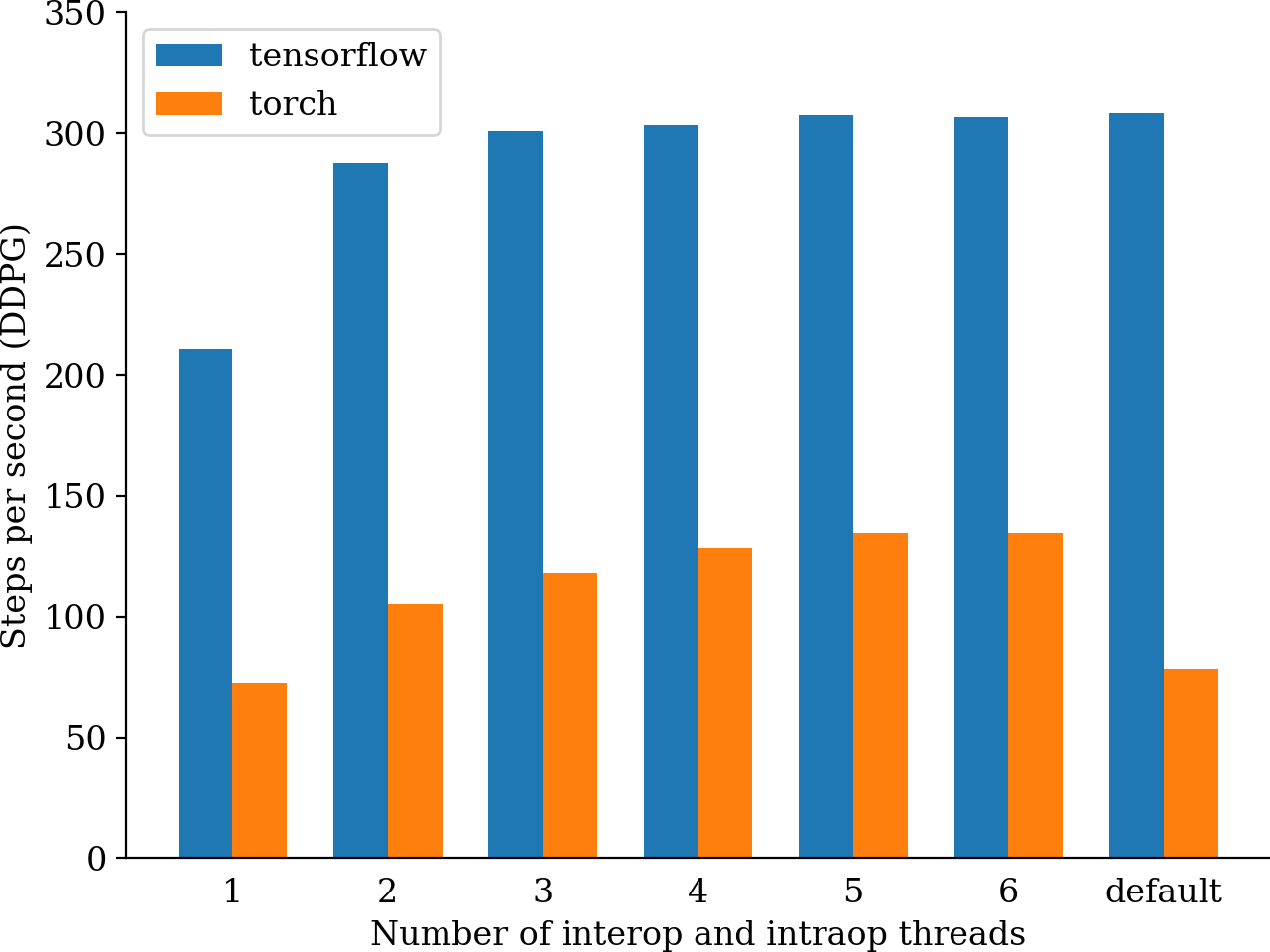}
    \caption{Speed comparison between TensorFlow 2 and PyTorch agents trained on walker-walk. Agents were trained for 1 million steps, using the same parameters as for the benchmark. The time spent to run the last 250,000 steps is used to measure the average number of steps per second indicated by bars. The agents were fully trained in turn on the same 6-core processor running at 3.8 GHz without GPU. The figure on the left used the default number of threads while the figure on right shows the impact of setting the number of interop and intraop threads for DDPG. The difference between the two frameworks might be due to a better optimization mechanism provided by TensorFlow's \lstinline[style=pythonstyle]{tf.function} decorator.}
    \label{figure:tf_vs_pt}
\end{figure}

\section{Comparison to Existing Libraries}

\begin{table}[h]
\caption{Comparison between Tonic and other popular existing RL libraries.}
\label{table:libraries}
\vskip 0.15in
\begin{center}
\begin{small}
\begin{sc}
\begin{tabular}{llllllll}
\toprule
Library            & Active   & Frameworks   & Modules & Trainer & Distributed     & Bench     & Scripts \\
\midrule
Tonic (ours)       & $\surd$  & TF 2 and PT  & $\surd$ & $\surd$ & sync with PEB   & large     & $\surd$ \\
RLlib              & $\surd$  & TF 1 and PT  &         & $\surd$ & sync and async  & small     & \\
Baselines          &          & TF 1         &         &         & some sync (MPI) &           & \\
Stable Baselines   &          & TF 1         &         &         & some sync (MPI) &           & \\
Stable Baselines 3 & $\surd$  & PT           &         &         &                 &           & \\
Spinning Up        &          & TF 1 and PT  &         &         & some sync (MPI) & small     & \\
Acme               & $\surd$  & TF 2 and JAX & $\surd$ & $\surd$ & async           &           & \\
RLgraph            &          & TF 1 and PT  & $\surd$ & $\surd$ & sync and async  &           & \\
Coach              &          & TF 1         & $\surd$ & $\surd$ & sync and async  & small     & \\
\bottomrule
\end{tabular}
\end{sc}
\end{small}
\end{center}
\vskip -0.1in
\end{table}

Repositories that have not received any major update during the past year are marked as not active. With MPI (Message Passing Interface) each worker has its own environment and a copy of the networks, computes gradients based on its own experience and a synchronous averaging of the gradients is required for each update. This approach does not easily scale to a large number of workers. Asynchronous distributed training does not guarantee reproducible experiments and provides significantly different results on machines with different compute power. Tonic is the only library properly handling timeout terminations and providing three essential scripts to train, plot and play easily.

\clearpage
\section{The Proposed TD4 Agent}

\subsection{Full Source Code}
\label{section:td4_code}

\begin{lstlisting}[style=pythonstyle,numbers=left]
import tensorflow as tf

from tonic import replays
from tonic.tensorflow import agents, models, normalizers, updaters


def default_model():
    return models.ActorTwinCriticWithTargets(
        actor=models.Actor(
            encoder=models.ObservationEncoder(),
            torso=models.MLP((256, 256), 'relu'),
            head=models.DeterministicPolicyHead()),
        critic=models.Critic(
            encoder=models.ObservationActionEncoder(),
            torso=models.MLP((256, 256), 'relu'),
            head=models.DistributionalValueHead(-150., 150., 51)),
        observation_normalizer=normalizers.MeanStd())


class TwinCriticDistributionalDeterministicQLearning:
    def __init__(
        self, optimizer=None, target_action_noise=None, gradient_clip=0
    ):
        self.optimizer = optimizer or \
            tf.keras.optimizers.Adam(lr=1e-3, epsilon=1e-8)
        self.target_action_noise = target_action_noise or \
            updaters.TargetActionNoise(scale=0.2, clip=0.5)
        self.gradient_clip = gradient_clip

    def initialize(self, model):
        self.model = model
        variables_1 = self.model.critic_1.trainable_variables
        variables_2 = self.model.critic_2.trainable_variables
        self.variables = variables_1 + variables_2

    @tf.function
    def __call__(
        self, observations, actions, next_observations, rewards, discounts
    ):
        next_actions = self.model.target_actor(next_observations)
        next_actions = self.target_action_noise(next_actions)
        next_value_distributions_1 = self.model.target_critic_1(
            next_observations, next_actions)
        next_value_distributions_2 = self.model.target_critic_2(
            next_observations, next_actions)

        values = next_value_distributions_1.values
        returns = rewards[:, None] + discounts[:, None] * values
        targets_1 = next_value_distributions_1.project(returns)
        targets_2 = next_value_distributions_2.project(returns)
        next_values_1 = next_value_distributions_1.mean()
        next_values_2 = next_value_distributions_2.mean()
        twin_next_values = tf.concat(
            [next_values_1[None], next_values_2[None]], axis=0)
        indices = tf.argmin(twin_next_values, axis=0, output_type=tf.int32)
        twin_targets = tf.concat([targets_1[None], targets_2[None]], axis=0)
        batch_size = tf.shape(observations)[0]
        indices = tf.stack([indices, tf.range(batch_size)], axis=-1)
        targets = tf.gather_nd(twin_targets, indices)

        with tf.GradientTape() as tape:
            value_distributions_1 = self.model.critic_1(observations, actions)
            losses_1 = tf.nn.softmax_cross_entropy_with_logits(
                logits=value_distributions_1.logits, labels=targets)
            value_distributions_2 = self.model.critic_2(observations, actions)
            losses_2 = tf.nn.softmax_cross_entropy_with_logits(
                logits=value_distributions_2.logits, labels=targets)
            loss = tf.reduce_mean(losses_1) + tf.reduce_mean(losses_2)

        gradients = tape.gradient(loss, self.variables)
        if self.gradient_clip > 0:
            gradients = tf.clip_by_global_norm(
                gradients, self.gradient_clip)[0]
        self.optimizer.apply_gradients(zip(gradients, self.variables))

        return dict(loss=loss)


class TD4(agents.TD3):
    def __init__(
        self, model=None, replay=None, exploration=None, actor_updater=None,
        critic_updater=None, delay_steps=2
    ):
        model = model or default_model()
        replay = replay or replays.Buffer(num_steps=5)
        actor_updater = actor_updater or \
            updaters.DistributionalDeterministicPolicyGradient()
        critic_updater = critic_updater or \
            TwinCriticDistributionalDeterministicQLearning()
        super().__init__(
            model=model, replay=replay, exploration=exploration,
            actor_updater=actor_updater, critic_updater=critic_updater,
            delay_steps=delay_steps)
\end{lstlisting}

\subsection{Command Lines}
\label{section:td4_commands}

Training can be started using:

\begin{lstlisting}[style=bashstyle]
python3 -m tonic.train \
--header "import td4, tonic.tensorflow" \
--environment "tonic.environments.ControlSuite('humanoid-walk')" \
--agent "td4.TD4()" \
--seed 0
\end{lstlisting}

The agent can be evaluated against D4PG and TD3 using:

\begin{lstlisting}[style=bashstyle]
python3 -m tonic.plot --path humanoid-walk --baselines D4PG TD3
\end{lstlisting}

\end{document}